# Learning Mixtures of Submodular Shells with Application to Document Summarization

**Hui Lin**
University of Washington
hlin@ee.washington.edu

**Jeff Bilmes**
University of Washington
bilmes@ee.washington.edu

## Abstract

We introduce a method to learn a mixture of submodular "shells" in a large-margin setting. A submodular shell is an abstract submodular function that can be instantiated with a ground set and a set of parameters to produce a submodular function. A mixture of such shells can then also be so instantiated to produce a more complex submodular function. What our algorithm learns are the mixture weights over such shells. We provide a risk bound guarantee when learning in a large-margin structured-prediction setting using a projected subgradient method when only approximate submodular optimization is possible (such as with submodular function maximization). We apply this method to the problem of multi-document summarization and produce the best results reported so far on the widely used NIST DUC-05 through DUC-07 document summarization corpora.

## 1 Introduction

Submodular functions [10] are those that satisfy the property of *diminishing returns*: given a finite ground set $V$, for any $A \subseteq B \subseteq V \setminus v$, a submodular function $f$ must satisfy $f(A \cup \{v\}) - f(A) \geq f(B \cup \{v\}) - f(B)$, i.e., the incremental "value" of $v$ decreases as the context in which $v$ is considered grows from $A$ to $B$. Submodular functions share a number of properties in common with convex and concave functions [29], including their wide applicability, their generality, their multiple options for their representation, and their closure under a number of common operators (including mixtures, truncation, complementation, and certain convolutions). For example, the weighted sum of a collection of submodular functions $\{f_i\}_i$, $f = \sum_i w_i f_i$ where $w_i$ are nonnegative weights, is also submodular. While they have a long history in operations research, game theory, econometrics, and electrical engineering, they are still beginning to be studied in machine learning for a variety of tasks including sensor placement [19, 20], structure learning of graphical models [34], document summarization [27, 28] and social networks [18].

The problem of learning submodular functions has also recently been addressed. For example, in [15], it is asked "can one make only polynomial number of queries to an unknown submodular function $f$ and constructs a $\hat{f}$ such that $\hat{f}(S) \leq f(S) \leq g(n)\hat{f}(S)$ where $n$ is the ground set size and for what function $g : \mathbb{N} \to \mathbb{R}$ is the possible?". Among many results, they show that even with adaptive queries and monotone functions, one cannot learn better than an $\Omega(\sqrt{n}/\log n)$ approximation of a given fixed submodular function. Similarly, [1] addressed the submodular function learning problem from a learning theory perspective, given a distribution on subsets. They provide strong negative results including that one can not approximate in this setting to within a constant factor. In general, therefore, learning submodular functions is hard.

While learning over all possible submodular functions may be hard, this does not preclude learning submodular functions with known forms with unknown parameters. For example, given a finite set of $M$ fixed submodular components $\{f_i\}_{i=1}^M$ where $f_i : 2^V \to \mathbb{R}$ is submodular over $V$, then learning a conical mixture $\sum_{i=1}^M w_i f_i$, where $(w_1, w_2, \ldots, w_M) = \boldsymbol{w} \in \mathbb{R}_+^M$, has not in the above been ruled out to have approximate guarantees. Such mixtures might span a very large set of submodular functions, depending on the diversity of the component set. We call such a problem "learning submodular mixtures."

In this paper we extend this one step further, to an approach we call learning "submodular shells." An instantiated submodular shell is a function $f_{\alpha,(V,\beta)} : 2^V \to \mathbb{R}$ indexed by a pair of parameter vectors $\alpha, \beta$ and a ground set $V$. The parameters $\beta$ are associated with a particular ground set $V$ but the parameters $\alpha$ of the shell apply to *any* ground-set vector pair $(V,\beta)$. A *shell* has the form $f_{\bar{\alpha},(\cdot,\cdot)}$, which does not have $(V,\beta)$ instantiated. We might learn, for example, that a particular value $\bar{\alpha}$ produces a good shell $f_{\bar{\alpha},(\cdot,\cdot)}$ for any instantiation of that shell with a particular $(V,\beta)$. In this sense, a submodular shell might be seen as a form of "structured submodularity." Moreover, we might learn $\boldsymbol{w}$ in a mixture of such shells $\sum_i w_i f_{\alpha_i,(\cdot,\cdot)}$ based on data consisting only of training tuples of the form $\{((V^{(t)}, \beta^{(t)}), S^{(t)})\}_t$ where $S^{(t)} \subseteq V^{(t)}$ and which can be used in an objective that involves the instantiated mixture of shells. Note that here, training tuples may consist of a sequence of different ground sets, and the goal is from these training tuples, and given a finite and fixed set of shells parameterized by $\{\alpha_i\}$, identify the conical weights $\boldsymbol{w}$ that optimize an objective. Of course, if all training ground sets are identical, then learning submodular shell mixtures reduces to learning submodular mixtures. In practice, however, the training ground sets are not usually identical. E.g., in extractive document summarization, the training data could be a set of documents and the corresponding human summaries, where the ground sets, those sentences that constitute the documents, are not identical between training samples. In particular, for a document $t$, the ground set $V^{(t)} = \{1, \cdots, n_t\}$ where $n_t$ is the number of sentences in document $t$, and $\beta^{(t)}$ could be the term-frequency vectors for the sentences (see Section 5.2 for more instances of this).

We introduce a method whereby submodular shell mixtures may be learnt in a max-margin structured-prediction setting, and provide a risk-bound guarantee under approximate submodular maximization (Theorem 1). We apply our method to the extractive document summarization problem and, as mentioned in the abstract, this yields extremely good results.

## 2 Structured Prediction

A main goal of learning is to identify a function $h : \mathcal{X} \to \mathcal{Y}$, that maps from an input domain $\mathcal{X}$ to an output domain $\mathcal{Y}$. Structured prediction problems [51, 50, 14] are those where these domains may consist of combinatorial structures. Often, given an input $\boldsymbol{x} \in \mathcal{X}$, only a subset of $\mathcal{Y}$, say $\mathcal{Y}_{\boldsymbol{x}}$, is valid. For example, if $\boldsymbol{x}$ is a sentence, and $\mathcal{Y}$ is the set of all parse trees, than only a subset of possible parse trees $\mathcal{Y}_{\boldsymbol{x}}$ might be valid. This offers little solace, however, as $\mathcal{Y}_{\boldsymbol{x}}$ is typically still exponentially large. Usually, one forms a score function $s : \mathcal{X} \times \mathcal{Y}$ that measures how good an output $\boldsymbol{y}$ is for a given input $\boldsymbol{x}$. The decision problem is, given an $\boldsymbol{x}$, find one of the outputs with the highest score, i.e., $\boldsymbol{y}^* \in \operatorname{argmax}_{\boldsymbol{y} \in \mathcal{Y}_{\boldsymbol{x}}} s(\boldsymbol{x}, \boldsymbol{y})$.

One of the primary challenges of structured prediction is that one needs to handle $\mathcal{Y}_{\boldsymbol{x}}$, whose cardinality is finite but exponentially large. A common way to address this issue is to make assumptions on both $\mathcal{Y}$ and $s$. For instance, if $s$ decomposes over the parts of $\mathcal{Y}$ and moreover only depends on "local" parts, one could employ dynamic programming or integer programming algorithm to find the optimal solution. Another approach is to explore combinatorial structures upon which efficient algorithms are available. Examples of this sort include the Hungarian method [21] for maximum weighted bipartite matching and the Chu-Liu-Edmonds algorithm [6, 9] for optimal branchings. Alternatively, as we do in the sequel, one could resort to approximate solutions with approximation guarantees.

## 3 Submodular Shell Scores

In this paper, we propose to explore the structured prediction problem using submodular shell score functions. That is, given $\boldsymbol{x}$, $s(\boldsymbol{x}, \cdot) : \mathcal{Y}_{\boldsymbol{x}} \to \mathbb{R}$ is a submodular function where $\mathcal{Y}_{\boldsymbol{x}}$ is considered to be the finite ground set associated with $\boldsymbol{x}$ — there is a finite ground set $V_{\boldsymbol{x}}$ associated with $\boldsymbol{x}$ where $\mathcal{Y}_{\boldsymbol{x}} \subseteq 2^{V_{\boldsymbol{x}}}$. There are at least two benefits of using such submodular score functions. First, submodular functions are natural and expressive with the capability of modeling decisions beyond local or linear interactions among parts. That is, submodular functions may allow for global direct interactions over a structured object unlike score functions that must decompose in some way. Second, this expressive power does require prohibitive computation as would an arbitrary score function. The reason is that the submodular maximization problem, even under many constrained settings, can be solved efficiently and near-optimally with rigorous performance guarantees [36, 13, 25, 2, 11].

The hypothesis function we consider takes the following linear discriminant form:

$$h(\boldsymbol{x};\boldsymbol{w}) = \operatorname*{argmax}_{\boldsymbol{y} \in \mathcal{Y}_{\boldsymbol{x}}} s(\boldsymbol{x},\boldsymbol{y}) = \operatorname*{argmax}_{Y \in \mathcal{Y}_{\boldsymbol{x}}} \boldsymbol{w}^\top \boldsymbol{f}_{\boldsymbol{x}}(Y)$$
$$= \operatorname*{argmax}_{Y \in \mathcal{Y}_{\boldsymbol{x}}} \sum_i w_i f_{\alpha_i,(V_{\boldsymbol{x}},\beta_{\boldsymbol{x}})}(Y)$$

where $\boldsymbol{w} \in \mathbb{R}_+$ is a non-negative weight vector, and $f_{\alpha,(V,\beta)} : 2^V \to \mathbb{R}$ is submodular over subsets of $V$ for every valid value of $\alpha$ and $(V,\beta)$. Since the weights are non-negative, the score function is also submodular. We call $f_{\alpha,(V,\beta)}$ a *submodular shell*, abstracting a set of submodular functions characterized by some $\alpha,\beta$ and a ground set $V$. The score function is called a *submodular shell mixture*, and each $f_{\alpha_i,(V,\beta)}$ is a *shell component* of the mixture. Note that the parameters $\beta$ are associated with a particular ground set $V$ but the parameters $\alpha$ of the shell apply to *any* pair of ground set $V$ and $\beta$ value.

While we discuss our learning process in Section 4, we introduce the learning problem here to improve clarity. We are given a set of training instances $S = \{(\boldsymbol{x}^{(t)}, \boldsymbol{y}^{(t)})\}_{t=1}^T$ drawn independently from a distribution $D$ over pairs $(\boldsymbol{x}, \boldsymbol{y}) \in \mathcal{X} \times \mathcal{Y}$ restricted in such a way that $\boldsymbol{y} \in \mathcal{Y}_{\boldsymbol{x}}$ for any pair $(\boldsymbol{x}, \boldsymbol{y})$. $\boldsymbol{x}^{(t)}$ corresponds to a finite ground set $V_{\boldsymbol{x}^{(t)}}$ and a set of parameters $\beta_{\boldsymbol{x}^{(t)}}$ and so we say that $\boldsymbol{x}^{(t)} = (V_{\boldsymbol{x}^{(t)}}, \beta_{\boldsymbol{x}^{(t)}})$. We also have a finite collection of $M$ submodular shells $\{f_{\alpha_i, (\cdot, \cdot)}\}_{i=1}^M$ each of which is instantiated into a submodular function for any $\boldsymbol{x} \in \mathcal{X}$. That is, $f_{\alpha_i, \boldsymbol{x}^{(t)}} : 2^{V_{\boldsymbol{x}^{(t)}}} \to \mathbb{R}$ is a submodular function. We are also given a loss function and $\ell_{\boldsymbol{x}, \boldsymbol{y}}(\hat{\boldsymbol{y}})$ that measures the loss of predicting $\hat{\boldsymbol{y}} \in \mathcal{Y}_{\boldsymbol{x}}$ when the true label is $\boldsymbol{y} \in \mathcal{Y}_{\boldsymbol{x}}$. The empirical risk minimization problem then is to find a conical mixture $\boldsymbol{w} \in \mathbb{R}_+^M$ such that the risk $\mathbb{E}_{(\boldsymbol{x}, \boldsymbol{y}) \sim D}[\ell_{\boldsymbol{x}, \boldsymbol{y}}(h(\boldsymbol{x}; \boldsymbol{w}))]$ is minimized. It is important to realize that what is learnt is the mixture coefficients $\boldsymbol{w}$ over a set of shells that may be instantiated into a weighted sum of submodular functions.

Before we discuss learning in detail, we consider the expressive power of mixtures of such submodular shells.

### 3.1 Submodular Shell Mixtures

A fairly rich subclass of submodular functions can be instantiated from submodular shell mixtures with components being simpler submodular shells. Consider, for example, truncation functions, mixtures of which can represent many submodular functions of interest. Given a modular function[1] $c : 2^V \to \mathbb{R}$, define a truncation function $t_{\alpha, (V, c)} : 2^V \to \mathbb{R}$ as

$$t_{\alpha, (V, c)}(A) = \min\left\{c(A), \alpha \max_{B \subseteq V} c(B)\right\},$$

which is submodular, where $\alpha \in \mathbb{R}$ is the truncation threshold. A modular function can also be written as a truncation function with $\alpha = \infty$. We note that $(V, \beta)$ above takes the form $(V, c)$ here, and we might have a collection of such functions $\{t_{\alpha_i, (V, c)}\}_i$, and learning a mixture $\boldsymbol{w}$ would mean that $w_i$ indicates the importance of truncation threshold $\alpha_i$.

In fact, many submodular functions can be written as mixtures of truncation functions. For example, coverage type functions, canonical examples of submodular functions, can be written as submodular mixtures. Precisely, let a collection of sets be $\{A_i\}_{i \in \{1, \cdots, |V|\}}$ on a ground set $E$. Then the set cover function becomes

$$f : 2^V \to \mathbb{R}_+, \quad f(B) = \left|\bigcup_{i \in B} A_i\right|.$$

We can define costs $c_{i,j} = 1$ if $A_i$ contains $j \in E$ and

---
[1] $f$ is modular if both $f$ and $-f$ are submodular.

$c_{i,j} = 0$ otherwise. We then have

$$f(B) = \left|\bigcup_{i \in B} A_i\right| = \sum_{j=1}^{|E|} \min\left\{\sum_{i \in B} c_{i,j}, 1\right\},$$

which is a mixture of truncation functions. Moreover sums of concave functions applied to cardinality functions [47] can be represented as mixtures of truncation functions — any sum of such functions can be expressed as a sum of a modular function and nonnegative linear combinations of truncation functions.

Another interesting class is weighted matroid rank functions [32]. Given matroids [56, 38] $\mathcal{M}_i = (V, \mathcal{I}_i)$, and modular functions $m_i : 2^V \to \mathbb{R}_+$, we get:

$$\sum_{i=1}^M w_i f^i_{m_i, (V, \mathcal{M}_i)} = \sum_{i=1}^M w_i \max\{m_i(I) : I \subseteq S, I \in \mathcal{I}_i\}$$

which includes cover-like functions and many others.

The submodular functions introduced in [28] for document summarization can also been seen as a mixture of submodular shells, with components being either the coverage or the diversity shell that can be instantiated for a particular document.

In general, by using rich enough families of shell components, a submodular shell mixture could be very expressive, representing a very large family of submodular functions. Moreover, another advantage of using submodular shell mixture representations is that, since we assume each component is given and only the component weights are unknown, the learning problem can be addressed by using well-established methods. And by learning shell mixtures, we can then apply the learnt mixture to structured problems even over different underlying ground sets.

## 4 Learning Submodular Shell Mixtures

While there might be many ways of learning shell mixtures, in this paper we take a large-margin approach, standard in structured prediction. In other words, we want to minimize the risk of making predictions (decisions) when using the submodular shell mixture as a score function. Of course, the standard maximization in learning structured prediction is only approximate in this case, since maximizing submodular functions in NP-hard (although constant-factor approximable). Moreover, we need to identify a valid loss function that dose not violate submodularity. We wish also to ensure that the learnt parameters have quality guarantees (e.g., a risk bound). All of this is done in the below.

## 4.1 Large Margin Learning

We consider the learning problem defined in Section 3. For concision, we denote

$$\mathcal{Y}_t \triangleq \mathcal{Y}_{\boldsymbol{x}^{(t)}}, \quad \boldsymbol{f}_t(\boldsymbol{y}) \triangleq \boldsymbol{f}_{\boldsymbol{x}^{(t)}}(\boldsymbol{y}), \text{ and } \ell_t(\boldsymbol{y}) \triangleq \ell_{\boldsymbol{x}^{(t)},\boldsymbol{y}^{(t)}}(\boldsymbol{y}).$$

We follow the maximum margin approach [51] to learn the component weights $\boldsymbol{w}$, where the goal is to find a score function that scores $\boldsymbol{y}^{(t)}$ higher than all other $\boldsymbol{y} \in \mathcal{Y} \setminus \boldsymbol{y}^{(t)}$ by some margin. Formally, the learning problem is as follows:

$$\min_{\boldsymbol{w} \geq 0} \frac{1}{T} \sum_{t=1}^{T} \hat{\ell}_t(\boldsymbol{w}) + \frac{\lambda}{2} \|\boldsymbol{w}\|^2, \tag{1}$$

where the generalized hinge loss

$$\hat{\ell}_t(\boldsymbol{w}) \triangleq \max_{\boldsymbol{y} \in \mathcal{Y}_t} \left(\boldsymbol{w}^\top \boldsymbol{f}_t(\boldsymbol{y}) + \ell_t(\boldsymbol{y})\right) - \boldsymbol{w}^\top \boldsymbol{f}_t(\boldsymbol{y}^{(t)}) \tag{2}$$

and a quadratic regularizer is minimized.

---
**Algorithm 1:** Projected subgradient descent for learning submodular shell mixtures.

---
**Input** : $S = \{(\boldsymbol{x}^{(t)}, \boldsymbol{y}^{(t)})\}_{t=1}^T$ and a learning rate sequence $\{\eta_t\}_{t=1}^T$.
$\boldsymbol{w}_0 = 0$;
**for** $t = 1, \cdots, T$ **do**
  Approximate loss augmented inference:
  $\hat{\boldsymbol{y}}_t \approx \text{argmax}_{\boldsymbol{y} \in \mathcal{Y}_t} \boldsymbol{w}_{t-1}^\top \boldsymbol{f}_t(\boldsymbol{y}) + \ell_t(\boldsymbol{y})$;
  Compute the subgradient:
  $\boldsymbol{g}_t = \lambda \boldsymbol{w}_{t-1} + \boldsymbol{f}_t(\hat{\boldsymbol{y}}_t) - \boldsymbol{f}_t(\boldsymbol{y}^{(t)})$;
  Update the weights with projection:
  $\boldsymbol{w}_t = \max(\boldsymbol{0}, \boldsymbol{w}_{t-1} - \eta_t \boldsymbol{g}_t)$;
**Return** : the averaged parameters $\frac{1}{T} \sum_t \boldsymbol{w}_t$.

---

Many algorithms have been proposed for the large margin learning problem, including those based on the exponentiated gradient method [7], the dual extragradient method [52], the cutting-plane algorithm [54], and the subgradient descent method [42, 43]. We adopt a subgradient descent algorithm to learn submodular shell mixtures, as illustrated in Algorithm 1, where max of two vectors takes dimension-wise maximum, i.e. $\max\{\boldsymbol{a}, \boldsymbol{b}\} = (\max(a_1, b_1), \cdots, \max(a_n, b_n))$ where $\boldsymbol{a}, \boldsymbol{b} \in \mathbb{R}^n$.

Note, to preserve submodularity, the component weights must be non-negative. We thus simply project the weights to the non-negative orthant whenever doing the updates, and it is easy to show that updates followed by projection onto a non-negative region do not affect the convergence or correctness of the algorithm as when a point is projected back into the convex set, it is moved closer to every point in the set including the optimal points.

Second, whether the learning can be done efficiently depends on whether the so called *loss augmented inference* (LAI) problem,

$$\max_{\boldsymbol{y} \in \mathcal{Y}_t} \boldsymbol{w}^\top \boldsymbol{f}_t(\boldsymbol{y}) + \ell_t(\boldsymbol{y}), \tag{3}$$

can be solved efficiently. The LAI problem has a term that precisely matches the prediction problem whose parameters we are trying to learn but also has an additional term corresponding to the loss. Tractability of LAI therefore not only depends on the tractability of the prediction problem but also on the form of loss functions. When using submodular shell mixtures as the score function, we use approximate inference with a performance guarantee. Therefore, we must perform approximate inference for the LAI problem as well. We thus in general refer to this as approximate learning.

## 4.2 Approximate learning

Since some form of inference is a dominant subroutine in many learning algorithms for structured prediction, it is natural to use good approximate inference techniques to make the learning problem tractable. When learning submodular shell mixtures, we inevitably must use approximate learning since, by using a more expressive class of score functions that need not decompose, the inference problem is intractable to do exactly. Using approximate inference as a drop-in replacement for exact inference in learning, however, could mislead the learning algorithm and result in poorly learnt models. This is analyzed in [24], where it is pointed out that approximate learning could fail even with an approximate inference method having approximation guarantees. In general, therefore, it is problematic to assume that an arbitrary choice of approximate inference method will lead to useful results when the learning method expects exact feedback. Choosing compatible inference and learning procedures is therefore crucial.

In the following, we aim to leverage the approximation guarantees of submodular optimization such that the performance of approximate learning of submodular shell mixtures can be bounded in some way. One possible way of bounding is to investigate the degree to which we can approximate the parameters $\boldsymbol{w}$ that would be obtained by exact learning, since the parameters themselves offer little utility if good prediction cannot be made from them. Alternatively, we can focus on the quality of prediction obtained from an approximately learned model. In particular, we seek to bound the risk gap. That is, the difference between the expected loss of predictions from an approximate (but efficient) scheme and from exact (but intractable) methods.

### 4.3 Analysis

Hence, we must further analyze whether using good approximate inference will lead us to good approximate learning in our case. Before doing so, we note that there are two types of approximation inference algorithms, namely undergenerating and overgenerating approximations.

Consider a maximization problem

$$\max_{\boldsymbol{y} \in \mathcal{Y}} f(\boldsymbol{y}) \triangleq f^*.$$

Undergenerating approximation algorithm always find a solution $\boldsymbol{y} \in \mathcal{Y}$ such that $f(\boldsymbol{y}) \leq f^*$, while overgenerating approximation algorithm always find a solution $\boldsymbol{y} \in \bar{\mathcal{Y}} \supseteq \mathcal{Y}$ such that $f(\boldsymbol{y}) \geq f^*$. A greedy algorithm or the loopy belief propagation [40] are instances of undergenerating approximation algorithms. Relaxation methods, e.g. linear programming relaxation, are overgenerating approximation algorithms. Note that undergenerating algorithms usually produce solutions that are within the feasible region of the problem. Overgenerating algorithms, on the other hand, generate solutions that might lie outside this feasible region. Therefore, solutions found by overgenerating algorithms sometimes need to be mapped back to the feasible region (e.g., rounding of linear programming produced solutions) in order to produce a feasible solution, during which the approximation guarantee no longer holds in some cases [44]. For learning submodular shell mixtures, we are particularly interested in undergenerating algorithms since the greedy algorithm, one of the undergenerating algorithms, offers near-optimal solutions for submodular maximization under certain (e.g., cardinality, budget, and matroid) constraints [36, 13].

Generalization bounds for approximate learning with cutting-plane algorithms, with either undergenerating or overgenerating inferences, have been shown in [12]. For subgradient descent methods, generalization analysis is available, but only for overgenerating cases, in [30, 22]. As far as we know, no generalization analyses are available for approximate learning with undergenerating subgradient methods. We fill this gap and offer risk bounds for approximate learning with undergenerating subgradient methods.

We need two definitions before moving on.

**Definition 1** ($\rho$-approximate algorithm). *Given $f : \mathcal{Y} \to \mathbb{R}^+$ and a maximization problem $\max_{\boldsymbol{y} \in \mathcal{Y}} f(\boldsymbol{y}) \triangleq f^*$, we call an (undergenerating) algorithm a $\rho$-approximate algorithm if it finds a solution $\boldsymbol{y} \in \mathcal{Y}$ such that $f(\boldsymbol{y}) \geq \rho f^*$, where $0 \leq \rho \leq 1$.*

**Definition 2** ($\gamma$-approximate subgradient). *Given $f : \mathbb{R}^M \to \mathbb{R}$, a vector $\boldsymbol{g} \in \mathbb{R}^M$ is called a $\gamma$-subgradient of $f$ at $\boldsymbol{w}$ if for all $\boldsymbol{w}'$, $f(\boldsymbol{w}') \geq f(\boldsymbol{w}) + \boldsymbol{g}^\top(\boldsymbol{w}' - \boldsymbol{w}) - \gamma f(\boldsymbol{w})$ where $0 \leq \gamma \leq 1$ and $\boldsymbol{w}, \boldsymbol{w}' \in \mathbb{R}^M$.*

In Algorithm 1, one needs to solve the LAI in Eqn (3). Note that if $\ell$ is modular (e.g. hamming loss) or submodular (e.g. see Section 5.3), a $\rho$-approximate inference algorithm can also apply to this loss augmented inference to find a near-optimal solution efficiently. However, as mentioned above, when an approximate inference algorithm is used in a learning algorithm, a good approximation of the score might not be sufficient, and it is possible that the learning can fail even with rigorous approximate guarantees [24]. On the other hand, Ratliff et al. [43] show that the subgradient algorithm is robust under approximate settings, and the risk experienced during training with $\gamma$-approximate subgradients can be bounded.

Note that a $\rho$-approximate LAI does not necessary imply any $\gamma$-subgradient because the approximate ratio does not apply to the term $-\boldsymbol{w}^\top f_t(\boldsymbol{y}^{(t)})$. To analyze the actual impact of $\rho$-approximate LAI in the learning procedure when compared with the exact formulation, then, we provide risk bounds for the approximate learner in Theorem 1.

**Theorem 1.** *Assume $w_i, f_i, i = 1, \cdots, M$ are all upper-bounded by 1, $\hat{\ell}_t(\boldsymbol{w}) \leq B$, and $\|\boldsymbol{g}_t\| \leq G$. Let $\boldsymbol{w}^*$ and $\hat{\boldsymbol{w}}$ be the solutions returned by Algorithm 1 using exact and $\rho$-approximate LAI, respectively, with learning rate $\eta_t = \frac{2}{\lambda t}$ and $\lambda = \frac{G}{M}\sqrt{\frac{2(1+\log T)}{T}}$. Then for any $\delta > 0$ with probability at least $1 - \delta$,*

$$\mathbb{E}_{(\boldsymbol{x},\boldsymbol{y}) \sim D}[\ell_{\boldsymbol{y}}(h(\boldsymbol{x}; \hat{\boldsymbol{w}}))] \leq \frac{1}{\rho}\left(\frac{1}{T}\sum_{t=1}^T \hat{\ell}_t(\boldsymbol{w}^*)\right) + S(T),$$

*where*

$$S(T) = \frac{MG}{\rho}\sqrt{\frac{2(1+\log T)}{T}} + B\sqrt{\frac{2}{T}\log\frac{1}{\delta}} + \frac{1-\rho}{\rho}M.$$

The proof is given in the supplementary material. Note that there are three terms in $S(T)$. While the first two terms vanish as $T \to \infty$, the third term does not. Therefore, the additional risk incurred due to the use of $\rho$-approximate LAI for learning is $(R^* + (1-\rho)M)/\rho$, where $R^* = \lim_{T \to \infty} \frac{1}{T}\sum_{t=1}^T \hat{\ell}_t(\boldsymbol{w}^*)$ and can be seen as the risk of predictions using models learnt with exact inference. Thus, the better (larger $\rho$) the approximation, the less the additional risk there will be when using an approximate LAI. And when exact inference is used ($\rho = 1$), the additional risk shrinks to zero. Note that Theorem 1 applies to any loss function and any score function which is not necessary to be submodular. When addition assumptions are made (e.g., assumption that the loss function is linearly realizable as in [57]), a better bound might be possible where the additional risk shrinks to zero as $T$ grows. On the other hand, when the LAI objective is monotone submodular, a simple and efficient greedy algorithm performs near-optimally with approximation factor $1 - 1/e$ [36]. In practice, moreover, as the approximation factor of the greedy algorithm on submodular maximization is usually very close to 1 [27], one could expect very little additional risk when using Algorithm 1 with approximate inference on learning submodular shell mixtures.

To the best of our knowledge, Theorem 1 is the first approximate learning bound for subgradient algorithms with undergenerating (greedy) inference.

## 5 Application to Document Summarization

Submodular mixtures could be applied to many structured prediction problems of practical interest. In this paper, we apply submodular shell mixture learning to extractive document summarization as a case study.

### 5.1 Submodularity in Document Summarization

Extractive document summarization can be seen as a subset selection problem [27]. Given a ground set of sentences $V$, the task of extractive document summarization is selecting a subset of sentences, say $S$, that best represents the whole document. In other words, we want to find $A \subseteq V$ such that

$$A \in \underset{B \subseteq V}{\operatorname{argmax}} f(B) \text{ subject to: } \sum_{i \in B} c_i \leq b, \quad (4)$$

where $c_i \in \mathbb{R}^+$ is the cost of sentence $i$ (e.g., it could be the number of words in the sentence), $b \in \mathbb{R}^+$ is the total budget (e.g., it could the largest number of words allowed in a summary), and $f : 2^V \to \mathbb{R}$ is a set function that models the quality of a summary. Eqn (4) is known as the problem of submodular maximization subject to knapsack constraints [31] which NP-complete [35]. However, when $f$ is *monotone submodular*, Eqn (4) can be solved efficiently and near-optimally with a theoretical guarantee via greedy algorithms [48, 27].

One can always force $f$ to be submodular, leading to an objective function that can be optimized well but might on the other hand poorly represent a given problem. One attractive property of submodularity, like convexity in continuous domain, is that it arises naturally in many applications. One such applications is document summarization. As pointed out in [28], many well-established methods, including the widely used maximum margin relevance method [3], actually correspond to submodular optimization. Moreover, it is shown that the commonly used ROUGE score [26] for automatic summarization evaluation is monotone submodular [28], giving further evidence that submodular functions are natural for document summarization.

In this paper, we further show that not only is the ROUGE score submodular, the score used in the Pyramid method [37], one of the manual evaluation metrics that has been used in recent TAC summarization track[2], is also monotone submodular.

**Theorem 2.** *The modified score in Pyramid method is monotone submodular.*

The proof is in Appendix C in the supplement.

The remaining question is how to design (or ideally learn) a good submodular function for summarization. Lin and Bilmes [28] proposed a class of submodular functions that models the coverage as well as the diversity of summary. In this paper, we further generalize their class of submodular functions and propose to use *submodular shell mixtures* for document summarization.

### 5.2 Submodular shells for summarization

**Diversity shell components**

We define a diversity shell component as

$$f^{\text{diversity}}_{(a,K,\mathcal{A}),(V,\boldsymbol{r})}(S) = \frac{\sum_{k=1}^{K} \left(\sum_{i \in S \cap P_k} r_i\right)^a}{\sum_{k=1}^{K} \left(\sum_{i \in P_k} r_i\right)^a}, \quad (5)$$

where $0 \leq a \leq 1$ is the curvature, $K \in \mathbb{Z}^+$ is the number of clusters (partitions), $\mathcal{A}$ is a clustering algorithm, and $\{P_k\}_{k=1,\cdots,K}$ is a partition of the ground set $V$ generated by $\mathcal{A}$, and $\boldsymbol{r} = \{r_i\}_{i=1}^{|V|}$ with $r_i \in [0, 1]$ is the vector of singleton reward of element $i \in V$. The diversity component models the diversity of a summary set $S$, by diminishing the benefit of choosing elements from the same cluster.

Note that the $\alpha$ parameter of a submodular shell here takes the form $(a, K, \mathcal{A})$. By using different values of $a$ and $K$, and different clustering algorithms $\mathcal{A}$, we can produce a variety of submodular shells. The $(V, \beta)$ parameter of a submodular shell takes the form of $(V, \boldsymbol{r})$. When a document (ground set) is given, rewards of each sentence (i.e., $r_i$) can be computed, and the diversity shell component is then instantiated into a submodular function that measures the diversity of a summary for this particular document.

**Clustered facility location shell components**

We define clustered facility-location like components as

$$f^{\text{c-facility}}_{(K,\mathcal{A}),(V,\boldsymbol{r})}(S) = \frac{1}{K} \sum_{k=1}^{K} \max_{i \in S \cap P_k} r_i, \quad (6)$$

---

[2]http://www.nist.gov/tac/2011/Summarization/

where $K \in \mathbb{Z}^+$ is the number of clusters (partitions), $\mathcal{A}$ is a clustering algorithm, and $r_i \in [0, 1]$ is the singleton reward of element $i \in V$. This function has a similar form to the well known submodular facility location function, but defined on a partition of the ground set. We thus call it clustered facility location. If a summary contains multiple elements from a same cluster, the element with largest singleton reward will be regarded as the "representative" of this cluster, and only the reward of this representative will be counted into the final score. This again diminishes returns of choosing elements from the same cluster and therefore $f^{\text{c-facility}}$ is submodular.

**Fidelity shell components**

Given a ground set $V$, we define fidelity components as

$$f^{\text{fidelity}}_{\alpha, \left(V, \{C_i\}_{i=1}^{|V|}\right)}(S) = \frac{1}{|V|} \sum_{i \in V} \min\left\{\frac{C_i(S)}{C_i(V)}, \alpha\right\}, \quad (7)$$

where $0 < \alpha \leq 1$ is a saturation threshold and $C_i : 2^V \to \mathbb{R}$ is a *monotone* submodular function modeling how $S$ covers the information contained in $i$. This function is the normalized version of the coverage function defined in [28]. Basically, the saturation threshold controls how much of a given element $i \in V$ should be covered; once $C_i(S)$ is large enough such that the ratio of it over its largest possible value ($C_i(V)$) is above threshold, covering more of $i$ does not further increase the function value. Therefore, a larger value of $f^{\text{fidelity}}$ tends to have more $i \in V$ well covered. When a document is given, we can instantiate different submodular shells using a variety of $C_i$.

## 5.3 A Submodular Loss Function

The most widely used evaluation criteria for summarization is the ROUGE score, which is basically a submodular function that counts n-gram recall rate over human summaries. Let $S$ be the candidate summary (a set of sentences extracted from the ground set $V$), $c_e : 2^V \to \mathbb{Z}_+$ be the number of times n-gram $e$ occurs in summary $S$, and $R_i$ be the set of n-grams contained in the reference summary $i$ (suppose we have $K$ reference summaries, i.e., $i = 1, \cdots, K$). Then ROUGE-N [26] can be written as the following set function:

$$f_{\text{ROUGE-N}}(S) \triangleq \frac{\sum_{i=1}^K \sum_{e \in R_i} \min(c_e(S), r_{e,i})}{\sum_{i=1}^K \sum_{e \in R_i} r_{e,i}},$$

where $r_{e,i}$ is the number of times n-gram $e$ occurs in reference summary $i$. $f_{\text{ROUGE-N}}(S)$ is submodular, as shown in [28], but cannot be used as a loss function since it basically measures "accuracy" rather than loss.

An alternative is to use $1 - f_{\text{ROUGE-N}}(S)$ as a loss function, but this is supermodular. Note that in order to have the risk of the approximated learned model bounded, performance guarantees are required for the approximation algorithms used in loss augmented inference. When using $1 - f_{\text{ROUGE-N}}$, which is supermodular as a loss function, in the objective function for loss augmented inference (Eqn. (3)) along with a submodular shell mixture as the score function, the resulting objective function for LAI is then a submodular function plus a supermodular function. While an algorithm (e.g., submodular-supermodular procedure [33, 17]) is available to approximately optimize the sum of a submodular function and a supermodular function, performance guarantees usually do not exist for these algorithms (although this strategy might work well in practice and should ultimately be tested). Therefore, when using one-minus-ROUGE as the loss function, the greedy algorithm no longer provides a near-optimal solution when applied to the non-submodular objective, and the risk bound shown in Theorem 1 no longer holds.

To address this issue, we propose a ROUGE-like loss function that measures the "complement recall":

$$\ell_{\text{ROUGE}}(S) \triangleq \frac{\sum_{e \in \bar{R}} \omega_e c_e(S)}{\sum_{e \in \bar{R}} \omega_e r_e}, \quad (8)$$

where $\bar{R} = N \setminus \bigcup_i R_i$, and $N$ is the set of all the n-grams occur in the set of documents, and $r_e = c_e(V)$ is the number of times n-gram $e$ occurs in all the documents, $\omega_e$ is a non-negative weight for $e$, and $\bar{R}$ is the set of $n$-grams that are *not* covered by any human reference summary. Instead of counting with respect to a reference summary, $\ell_{\text{ROUGE}}$ counts the $n$-grams of a candidate summary $S$ w.r.t. the *complement* of reference summaries.

Intuitively, we want a summary $S$ to cover as many reference n-grams as possible so that it will get a high ROUGE-score; this is similar to having $S$ be large and overlapping as little as possible with the n-grams that are *not* in human references. In this sense, $\ell_{\text{ROUGE}}$ measures the portion of how many n-grams in the complement of the reference $n$-grams set are covered, and when comparing summaries with the same size, the smaller $\ell_{\text{ROUGE}}$ is, the better. The best case, i.e., the human reference itself, will have $\ell_{\text{ROUGE}}$ equal to 0.

Obviously, a poor summary that would also have $\ell_{\text{ROUGE}}$ equal to 0 is an empty summary. It is worth noting that $\ell_{\text{ROUGE}}$ only makes sense when comparing summaries that are close to the same budget. Fortunately, most summarization algorithms try to consume every bit of the budget in order to consume as much information as possible under the budget constraint. For summaries produced in this way, $\ell_{\text{ROUGE}}$ offers a fair indicator of their quality: the smaller the loss value, the larger the number of reference $n$-gram overlaps there are, and therefore the better the summary. We use the greedy algorithm for solving the summarization problems (Eqn. (4)). Due to its greedy nature, the algorithm always outputs summary candidates whose costs are close to the budget, and therefore $\ell_{\text{ROUGE}}$ can serve as a reasonable surrogate loss function for learning submodular shell mixtures for summarization tasks.

The major advantage of using $\ell_{\text{ROUGE}}$ as a loss function, of course, is algorithmic. The submodular learning analysis in Theorem 1 relies on the fact that a $\rho$-approximation algorithm is available for the loss augmented information. Since we use a submodular score function for summarization, the above objective will be submodular if the loss function is submodular. Fortunately, similar to $f_{\text{ROUGE-N}}$, the proposed loss for summarization, $\ell_{\text{ROUGE}}$, is also monotone submodular (in fact modular). Therefore, the LAI in submodular shell mixture learning for summarization is exactly the budgeted submodular maximization problem (Eqn (4)), and efficient and near-optimal algorithms are then available. Consequently, all the theoretical analyses in Section 4.3 apply.

We will soon see (Section 7) that using the ROUGE-like loss function proposed here empirically outperforms using $1 - f_{\text{ROUGE}}$ as a loss functions.

## 6 Related work

Recently, Yue and Guestrin [57] studied a linear submodular bandits problem in an online learning setting for optimizing a general class of feature-rich submodular utility models in diversified retrieval, and their theoretical result is based on the assumption that the award function has a special (linear) form. Raman et al. [41] also propose an online learning model and algorithm for learning rankings that exploiting feedback to maximize any submodular utility measure. While our approach and analysis could also be applied to the online setting, the focus and analysis in this paper are more on the batch learning. The submodular utility models in [57, 41], however, can both be seen as special instances of submodular shell mixtures.

More closely related to our work is that of [46] where large-margin learning of submodular score functions for extractive document summarization is studied. The representation of a submodular score function in [46], again, turns out to be a special case of a submodular shell mixture. Moreover, our submodular shell mixture framework is more general and flexible than the framework proposed in [46]. Although the authors claim that their approach applies to all submodular summarization models, there are many submodular functions useful for summarization that are not linear on either pairwise similarity or singleton importance. For example, in the diversity reward function, we have a concave function over the sums of singleton rewards, and even if using linear model for singleton rewards, the score function is non-linear over parameters since the weights can not be linearly extracted, and thus the algorithms in [46] do not apply. Viewing each feature as input to a submodular component, on the other hand, preserves the linearity on the parameters, whereas Algorithm 1 applies. Moreover, representing a score function using a submodular shell mixture is expressive, as we have shown in Section 3.1.

In [46], a cutting plane algorithm with one-minus-ROUGE F-measure as loss function was used to learn model weights. When doing the loss augmented inference, they use the greedy algorithm introduced in [27] to approximately optimize the objective. Their objective function, however, is not submodular, due to the non-submodular loss function they use. Therefore, the LAI inference is no longer guaranteed to be near-optimal, and the performance of approximate learning with their cutting plane algorithm is no longer guaranteed [12]. Also, Sipos et al. [46] apparently neglect the fact that the learned similarity (or importance) should be non-negative, which is a necessary ingredient to preserve the submodularity of the learned score function. One simple way to ensure this is to constrain the weights to be non-negative, as we do for submodular shell mixtures by project the weights to the non-negative orthant (Algorithm 1).

## 7 Experiments

We evaluated our approach on NIST's DUC[3] data 2003-2007, and demonstrate results on both generic and query-focused summarization. DUC data were created by national institution of standard technology (NIST). The DUC evaluation is one of the standardized benchmark evaluations for document summarization and researchers continue to publish results on these data sets.

### 7.1 Query-independent summarization

Table 1: ROUGE-1 recall (R) and F-measure (F) results (%) on DUC-04. DUC-03 was used as development set.

| DUC-04 | R | F |
|---|---|---|
| Takamura and Okumura [49] | 38.50 | - |
| Wang et al. [55] | 39.07 | - |
| Lin and Bilmes [27] | - | 38.39 |
| Lin and Bilmes [28] | 39.35 | 38.90 |
| Kulesza and Taskar [23] | 38.71 | 38.27 |
| Best system in DUC-04 (peer 65) | 38.28 | 37.94 |
| Submodular Shell Mixture | **40.43** | **39.78** |

The summarization tasks in DUC-03 and DUC-04 are

---
[3]http://duc.nist.gov/

generic summarization tasks. We used DUC-03 as the training set for submodular shell mixture learning. There are in total 60 document clusters in the DUC-03 task, therefore we have 60 training examples in total. We used a submodular shell mixture with 15 fidelity components for this task. In particular, the $\mathcal{C}_i$ function we used is $\mathcal{C}_i(S) = \sum_{j \in V} \delta_{i,j}$, where $\delta_{i,j}$ is the pairwise sentence similarity between sentence $i$ and $j$. We used three types of sentence similarities. The first two are cosine similarities with unigram and bigram TF-IDF vectors respectively. The third similarity is again cosine similarity but on the vector generated by latent semantic analysis. For each of the similarity measures, we use five different saturation thresholds ($\alpha = 0.01, \cdots, 0.05$), and thus we have 15 fidelity components in total. We used Algorithm 1 with the ROUGE-like loss (Eqn. (8)) with $\omega_e = 1$ for all $e$ to learn this submodular shell mixture. The ROUGE-1 results are shown in Table 1. As we can see, the result of the learned submodular shell mixture significantly outperforms all other previous reported results. Note that the experiment in [46] used a non-standard setup where DUC-04 data were divided into training, development, and test sets with only 5 documents in their test set. Therefore, results reported in [46] are not directly comparable to other results (e.g., our results) that follow standard DUC evaluation setup.

### 7.2 Query-focused summarization

Table 2: ROUGE-2 recall (R) and F-measure (F) results on DUC-05 (%). We used DUC-06 and DUC-07 as training sets.

| DUC-05 | R | F |
|---|---|---|
| Daumé III and Marcu [8] | 6.98 | - |
| Lin and Bilmes [28] | 7.82 | 7.72 |
| Best system in DUC-05 (peer 15) | 7.44 | 7.43 |
| Submodular Shell Mixture | **8.44** | **8.39** |

Table 3: ROUGE-2 recall (R) and F-measure (F) results (%) on DUC-06, where DUC-05 and DUC-07 were used as training sets.

| DUC-06 | R | F |
|---|---|---|
| Celikyilmaz and Hakkani-tür [4] | 9.10 | - |
| Shen and Li [45] | 9.30 | - |
| Lin and Bilmes [28] | 9.75 | 9.77 |
| Best system in DUC-06 (peer 24) | 9.51 | 9.51 |
| Submodular Shell Mixture | **9.92** | **9.93** |

Since 2004, DUC summarization evaluations have concentrated on query-focused summarization. We tested our approach for summarization on DUC-05, DUC-06 and DUC-07 data. In particular, we used DUC-06,07 as the training set for the DUC-05 task, DUC-05,07 as the training set for the DUC-06 task, and DUC-05,06 as the training set for the DUC-07 task.

Table 4: ROUGE-2 recall (R) and F-measure (F) results (%) on DUC-07. DUC-05 and DUC-06 were used as training sets. Note that the peer 15 system in DUC-07 used commercial web search engine to expand the queries.

| DUC-07 | R | F |
|---|---|---|
| Toutanova et al. [53] | 11.89 | 11.89 |
| Haghighi and Vanderwende [16] | 11.80 | - |
| Celikyilmaz and Hakkani-tür [4] | 11.40 | - |
| Lin and Bilmes [28] | 12.38 | 12.33 |
| Best system in DUC-07 (peer 15) | 12.45 | 12.29 |
| Submodular Shell Mixture | **12.51** | **12.40** |

We used diversity components, clustered facility location components, and fidelity components to form the submodular shell mixture for query-focused summarization. Three clusterings with different numbers of clusters were created ($K = 0.1|V|, 0.2|V|, 0.3|V|$). As for the singleton rewards, we used both query-independent and query-dependent singleton rewards. The query-independent reward for $i$ is simply the summation of the pairwise similarities of other elements in $V$ to $i$. For the query-dependent reward, we simply used the number of terms (up to a bi-gram) that sentence $j$ overlaps the query $Q$, where the IDF weighting is not used (i.e., every term in the query, after stop word removal, was treated as equally important). Therefore, in total, we have 6 clustered facility location components. We further used three curvatures in diversity components ($\alpha = 0.5, 0.6, 0.7$), which gives 18 diversity components in total. With one additional fidelity component, we have 25 components in total. Compared to other results reported in the literature (Table 2, Table 3 and Table 4), as far as we know, our approach achieves the best results reported so far on DUC-05, DUC-06, and DUC-07.

## 8 Conclusions

In this paper, we propose the notion of learning submodular shells, which abstract a set of submodular functions that can be instantiated into submodular functions given the input of a structured prediction task. Given a set of training instances, we use subgradient descent to learn the mixture coefficients over a set of submodular shells that may be instantiated into a weighted sum of submodular functions. We show that the submodular shell mixture is very expressive, and the risk of learning can be bounded when only approximate inference in possible. When applied to the task of document summarization, our approach achieves the best results reported so far on standardized benchmark takes for query-focused extractive summarization tasks.